\documentclass[conference]{IEEEtran}

\ifCLASSINFOpdf

\else

\fi

\usepackage[cmex10]{amsmath}
\usepackage{amsthm}
\usepackage{amssymb}
\usepackage{mathrsfs}
\usepackage{graphicx}
\usepackage{float}
\usepackage{array}
\usepackage{epstopdf}
\usepackage{multirow}
\newcommand{\argmin}{\arg\!\min}
\newcommand{\argmax}{\arg\!\max}
\allowdisplaybreaks[1]	

\begin{document}
\title{Multispectral Palmprint Recognition Using Textural Features}

\author{\IEEEauthorblockN{Shervin Minaee}
\IEEEauthorblockA{ECE Department\\ 
New York University\\New York, USA\\
shervin.minaee@nyu.edu\\}
\and
\IEEEauthorblockN{AmirAli Abdolrashidi}
\IEEEauthorblockA{ECE Department\\ 
New York University\\New York, USA\\
abdolrashidi@nyu.edu\\}
}

\maketitle

\begin{abstract}

In order to utilize identification to the best extent, we need robust and fast algorithms and systems to process the data. Having palmprint as a reliable and unique characteristic of every person, we extract and use its features based on its geometry, lines and angles.
There are countless ways to define measures for the recognition task. To analyze a new point of view, we extracted textural features and used them for palmprint recognition. Co-occurrence matrix can be used for textural feature extraction. As classifiers, we have used the minimum distance classifier (MDC) and the weighted majority voting system (WMV). The proposed method is tested on a well-known multispectral palmprint dataset of 6000 samples and an accuracy rate of 99.96-100\% is obtained for most scenarios which outperforms all previous works in multispectral palmprint recognition.
\end{abstract}


\IEEEpeerreviewmaketitle
\IEEEoverridecommandlockouts

\section{Introduction}
\IEEEPARstart{T}{here} are many reasons to use identification; to make sure that the person about to receive information or rights is indeed the right one. Several ways of identification include keys, photographs, passwords and biological samples. Many reasons necessitate the use of biometric characteristics of a person in their identification, including uniqueness, reliability and difficulty to forge. That identification can serve in personalized or secured applications or both. Other methods are subject to being lost, forgotten, stolen or replicated without authorization and their purpose is defeated rather easier.

Not surprisingly, there also exist many ways of identification based on biometric data such as fingerprints \cite{1}, iris patterns \cite{2}, face \cite{3} and palmprints \cite{4}. Among these, palmprints are simpler in the sense of acquisition and do not change over time significantly. However they can be temporarily or permanently altered due to external factors such as burns or scars. The key for their recognition is to extract the features of every person out of the prominent lines and wrinkles on their palms.  Being a popular area of research, there are many sets of features and different approaches used for palmprint recognition \cite{4}. General approaches for palmprint recognition are either transforming palmprints into another domain, namely transform-based approaches, or extracting principal lines and wrinkles and other geometrical characteristics as distinguishing factors.

Of the many researches in this area, a portion is based on transform domain features; for example, in \cite{5}, Wu proposed to use wavelet energy as features; and Kong implemented a system that uses Gabor-based features for palmprint recognition \cite{6}.
There are also quite a few line-based approaches, since palm lines capture the unique characteristics of a palmprint. Jia \cite{8} used robust line orientation code for palmprint verification. Chen \cite{7} extracted creases from palms in a way that does not need any translations or rotations afterwards, and used them for palmprint matching. Some of the approaches use the palmprint information both in spatial and frequency domains. As an example, in \cite{9}, Minaee developed a multispectral pamlprint recognition program using both statistical and wavelet features and achieved a much higher accuracy rate than all the previous works in multispectral palmprint recognition.
Also in \cite{10}, Xu sought to utilize quaternion principal component analysis for multispectral palmprint recognition which also resulted in a high accuracy.

Here we follow a new approach for palmprint recognition. We use textural features which are extracted from the co-occurrence matrix of every block whose concept will be elaborated in the next sections. It can incorporate adjacent blocks into the computations as well, sensing the overall texture. To test them, we have used one of the most popular multispectral palmprint datasets available, created by the Polytechnic University of Hong Kong (PolyU) \cite{Database}. It includes a set of 12 palmprint samples from 500 people taken in two days under four distinct light spectra: red, green, blue and infrared. Multispectral methods require different samples of the same object in order to make a better decision. In this paper, it is assumed that in the image acquisition section, four images of each palm sample are captured using CCDs. These images are preprocessed and their most useful sections are cropped and extracted as regions of interest (ROI). For every spectrum, features are denoted by $F_j^{(r)}$, $F_j^{(g)}$, $F_j^{(b)}$ and $F_j^{(i)}$ respectively. Three different palmprint samples from the used dataset are shown in Figure 1.
\begin{figure}[1 h]
\begin{center}
    \includegraphics [scale=0.7] {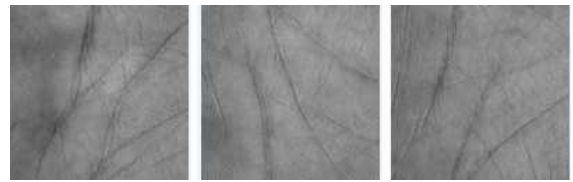}
\end{center}
  \caption{Three sample palmprints}
\end{figure}

After feature extraction, we have to use a classification algorithm to identify palmprints. In this work, the two employed methods of classification are minimum distance and weighted majority voting classifiers.

In this paper, the distribution of the contents is as follows; Section \ref{SectionII} provides a detailed explanation of the features and how to extract them; the weighted majority voting algorithm and minimum distance classifier are explained in Section \ref{SectionIII}; and the results of the experiments and comparisons are given in Section \ref{SectionIV}.

\section{Features}
\label{SectionII}

Features are an inevitable part of machine learning. The more informative features we have, the greater accuracy we get. Therefore for any classification or regression algorithm, it is particularly essential to extract the right set of features. Feature extraction algorithms have many applications in computer vision and object detection area. The most important step in image classification is that of defining a set of meaningful features to describe the pictorial information from the image blocks. Once these features are extracted, categorization can be executed using any classification technique.

For palmprint recognition, features come from various origins and types with different advantages and disadvantages. Statistical features like mean and deviation of pixels are common. Another popular class of features are transform-based including Fourier-, Gabor- and wavelet-based. Spatial and geometrical features can also prove efficient, especially in medical applications as illustrated by \cite{13} regarding chromosome segmentation. 

In human interpretation of color photographs, textural, spectral and contextual features are the three fundamental pattern elements. Textural features contain the spatial information of intensity variation in a single band. Spectral features describe the average intensity variation in different spectral bands. Contextual features contain information derived from neighboring regions of the area being analyzed \cite{14}.
Here we have extracted a set of textural features which are based on an outstanding work published in 1973 \cite{14}, in which the author introduced a general procedure for extracting textural properties of blocks of image data. These features are calculated in the spatial domain, and the statistical nature of texture is taken into account in this procedure, which is based on the assumption that the texture information in an image is contained in the overall or the ``average'' spatial relationship that the gray tones in the image have to one another.

To extract the features, each image is divided into non-overlapping blocks of size $N\times N$, the co-occurrence matrix for each block is constructed, and finally 14 features will be extracted from it. These features contain information about textural characteristics of such image, such as homogeneity, gray-tone linear dependencies and structure, contrast, number and nature of boundaries present and the whole complexity of the image.

Before further exploring the details of feature extraction, it is noteworthy how the proper block size should be chosen. If the chosen block size is too small, it will not have enough textural information to discriminate images of different people, while if too large a block size is selected, that block may have patterns belonging to different categories. Therefore the right block size should not be in the extremes. Here $N=16$ is chosen by trial and error based on the images in the dataset.

\subsection{Co-occurrence Matrix}

As the name suggests, co-occurrence matrix is a matrix defined over an image to measure the distribution of co-occurring intensity values for a given offset. We can denote any image as a two-dimensional function which maps any pair of coordinate to an intensity value, i.e., $I: X\times Y \rightarrow G$, where $X=\{1,2,3,...,N_x\}$ and $Y=\{1,2,3,...,N_y\}$ denote the horizontal and vertical spatial domains respectively, and $G$ denotes the set of all possible grayscale levels. For most images,  $G=\{0,1,2,...,255\}$. 

Then the co-occurrence matrix $P$ of image $I$ with the offset $(\Delta_x,\Delta_y)$ can be defined as:
\begin{gather*}
P_{\Delta_x,\Delta_y}(i,j)= \sum_{m=1}^{N_x} \sum_{n=1}^{N_y} \delta(I(m,n)-i)~ \delta(I(m+\Delta_x,n+\Delta_y)-j)
\end{gather*}
where $\delta(x)$ denotes the discrete Dirac function, which is 1 when the argument is zero, and 0 elsewhere. Therefore $P_{\Delta_x,\Delta_y}(i,j)$ counts how many times two pixels with intensities $i$ and $j$ are located in a distance of $(\Delta_x,\Delta_y)$ from each other. The offset $(\Delta_x,\Delta_y)$ depends on the direction $\theta$. Here we have used $(\Delta_x,\Delta_y)=(1,0)$. The neighborhood direction $\theta$ can be defined accordingly:
\begin{gather*}
\theta= tan^{-1}(\frac{\Delta_y}{\Delta_x})
\end{gather*}
It should be noted that the co-occurrence matrix has a size of ${N_g} \times {N_g}$, where $N_g$ denotes the number of gray-levels in the image. Here we quantized our images with a quantization step-size of 8, hence $N_g=32$.\\
As an example, consider the image matrix A as:
\begin{gather*}
A =
 \begin{pmatrix}
  1 & 1 & 2 & 1 \\
  2 & 3 & 1 & 2 \\
  2 & 1 & 3 & 2  \\
  3 & 3 & 2 & 1
 \end{pmatrix}
\end{gather*}
 Here A has only three different grayscale levels. Therefore its co-occurrence matrix has a size of 3$\times$3. The co-occurrence matrix of A for $(\Delta_x,\Delta_y)=(1,0)$ will be:
 \begin{gather*}
C=
 \begin{pmatrix}
  1 & 2 & 1\\
  3 & 0 & 1 \\
  1 & 2 & 1 
 \end{pmatrix}
\end{gather*}
Here, for example $C(1,2)$ counts how may times the cases $A(i,j)=1$ and $A(i+1,j)=2$ occur in matrix $A$, which is twice.

\subsection{Textural Feature Extraction From Co-occurrence Matrix }
After the co-occurrence matrix has been extracted, the following 14 textural features for each block may be extracted with ease. These features, which are described below, are similar to those in \cite{14}. For notation brevity, we first define the following terms derived from the matrix which will be used in the definition of the used features:
\begin{align*}
&p(i,j)= P(i,j)/R, \ \ \     \text{Normalized Co-occurrence Matrix }\\
&p_x(i)= \sum_{j=1}^{N_g}p(i,j) \ \ \  \ \    \text{Marginal Probability }\\
&p_y(j)= \sum_{i=1}^{N_g}p(i,j) \ \ \  \ \    \text{Marginal Probability }\\
&p_{x+y}(k)= \underset{i+j=k}{\sum \sum }p(i,j), \ \ \ k= 2,3,...,2N_g. \\
&p_{x-y}(k)= \underset{|i-j|=k}{\sum \sum}p(i,j), \ \ \ k= 0,1,...,N_g-1. \\
&HXY= -\sum_{i}\sum_{j} p(i,j) \log\big(p(i,j)\big) \\
&HXY1= -\sum_{i}\sum_{j} p(i,j) \log\big(p_x(i)p_y(j)\big) \\
&HXY2= -\sum_{i}\sum_{j} p_x(i)p_y(j) \log\big(p_x(i)p_y(j)\big) \\
&Q(i,j)= \sum_{k} \frac{p(i,k)p(j,k)}{p_x(i)p_y(k)}
\end{align*}
Now we define the following 14 textural features using these terms:
\begin{align*}
&f_1= \sum_{i}\sum_{j} \big[p(i,j) \big]^2,  \ \ \ \ \ \ \  \ \ \ \ \ \ \text{Angular Second Moment} \\
&f_2= \sum_{k=0}^{N_g-1}k^2 p_{x-y}(k), \  \ \ \ \ \ \ \ \ \ \ \  \ \ \text{Contrast} \\
&f_3= \frac{\sum_{i}\sum_{j}ijp(i,j)-\mu_x\mu_y}{\sigma_x\sigma_y}, \ \ \ \ \text{Correlation} \\
&f_4= \sum_{i}\sum_{j} (i-\mu)^2p(i,j), \ \ \ \ \ \ \ \text{Variance} \\
&f_5= \sum_{i}\sum_{j}  \frac{1}{1+(i-j)^2} p(i,j), \ \ \text{Inverse Diference Moment}\\
&f_6= \sum_{k=2}^{2N_g}k p_{x+y}(k), \ \ \ \ \ \ \ \ \ \  \ \ \ \ \ \ \  \text{Sum Average} \\
&f_7= \sum_{k=2}^{2N_g}(k-f_6)^2 p_{x+y}(k), \  \ \ \ \ \ \  \ \text{Sum Variance} \\
&f_8= -\sum_{k=2}^{2N_g} p_{x+y}(k) \log(p_{x+y}(k)), \ \text{Sum Entropy} \\
&f_9= -\sum_{i}\sum_{j} p(i,j) \log(p(i,j)), \   \text{Entropy} \ \\
&f_{10}= \sum_{k=0}^{N_g-1}(k-\mu_{x-y})^2 p_{x-y}(k),  \ \ \text{Diference Variance} \\
&f_{11}= -\sum_{k=0}^{N_g-1} p_{x-y}(k) \log\big( p_{x-y}(k) \big), \ \text{Difference Entropy} \\
&f_{12}= \frac{HXY-HXY1}{max\{HX,HY\}}  \\~\\
&f_{13}= \sqrt{1-exp{[-2(HXY2-HXY)]}}  \\~\\
&f_{14}= \sqrt{\text{Second largest eigenvalue of} \ Q}  \\
\end{align*}

Here $\mu_x$ and $\sigma_x$ denote the mean and standard deviation of the marginal distribution $P_x$ respectively. The same applies to $\mu_y$ and $\sigma_y$.

In the original paper, these 14 features have been defined, but it is suggested to calculate them for 4 angular co-occurrence matrices and take the average and range of each feature as a new feature, resulting in 28 features to be used. Here we use the 14 features for $\theta=0$.
The feature vector can be denoted as  $\textbf{f}=(f_1,f_2,...,f_{14})^\intercal$. It is necessary to find the mentioned features for each block of a palmprint. If each palm image has a size of $s_1 \times s_2$, the total number of non-overlapping blocks will be:
\begin{gather*}
M=\frac{s_1s_2}{N^2}
\end{gather*}
Therefore there are $M$ such feature vectors. Similarly, they can be put in the columns of a 2-dimensional matrix to produce the feature matrix of that palmprint, $\textbf{F}$:
\begin{gather*}
\textbf{F}=[\textbf{f}^{(1)} ; \textbf{f}^{(2)}; ... ;\textbf{f}^{(M)}]
\end{gather*}
Therefore there will be $14\times\textbf{M}$ features for each image (Here $\textbf{M}=64$).

\section{Recognition algorithm}
\label{SectionIII}
 After capturing the features of all people, a classifier should be used to compare the features of each test palmprint to all the training samples available and find its closest match. In this paper, two different classifiers are employed for this task. Weighted majority voting is inspired by counting votes from the features to the subjects. The minimum distance classifier, on the other hand, finds the minimum distance between the feature matrices of the training samples and test subjects. They are both explained in this section. Our only objective is to minimize the recognition error for the test samples, but when using a small database, issues such as the over-fitting problem should also be taken into account \cite{15}.

\subsection{Weighted Majority Voting}
\label{SubsectionIIIA}

In voting, there are referees that decide the answer by themselves and their votes are taken into account based on their importance, or weight. This scheme is very popular in learning algorithms and artificial intelligence. Unweighted voting is when we know all features should have the same effect on the outcome, but usually, each feature should use a different weight, either fixed or adaptive. When added, the total score will decide which person is the owner of the test image. Here the voters are the used features and they are weighted in a fixed manner. Apart from its simplicity, it also takes little time.

First, the images of every single person are randomly rearranged in the database so that the training part can use uniform data from all of the set of the 12 images. The training features are then collected, averaged and stored. Later, the other images are used as test subjects and the distance between the average feature matrix and that of every subject is measured. The case with the least distance with the subject is given points equal to the weight of the feature. In the end, the person gaining the maximum points is the winner. 

For every $\textbf{f}_i^{(t)}$, the voting result is: 
$$k^*(i)=\argmin_k||\textbf{f}_i^{(t)}-\textbf{f}_i^{(k)}||_2$$
When $\textbf{f}_i$ finds the person with minimum distance to the test subject, that person receives a point.
If the score of person $j$ based on $\textbf{f}_i$ is denoted by $w_iS_j(i)$ or $w_iI(j=\argmin_k|\textbf{f}_i^{(t)}-\textbf{f}_i^{k}|)$, where $w_i$ is the weight of the feature $i$ and $I(x)$ is an indicator function, the total score of the $j$-th training sample based on all the features in the scope of all the colors can be computed.
$$S_j=\sum_{All~colors}\sum_{i=1}^{i_{max}}{w_iI(j=\argmin_k|\textbf{f}_i^{(t)}-\textbf{f}_i^{k}|)}$$

In the end, the identification factor $j^*$ is:
\begin{gather*} 
j^*=\argmax_j \big[ S_j\big]= \argmax_j \big[\sum_{All~colors}\sum_{i}w_iS_j(i) \big]
\end{gather*}

\subsection{Minimum Distance Classifier}
\label{SubsectionIIIB}
The minimum distance classification is quite popular in the template matching area. It finds the distance between the features of the training samples and those of an unknown subject, and picks the training sample with the minimum distance to the unknown as the answer. To put it in equation, if we show the features of the test subject as $F^*$ and those of the test sample $i$ with $F^{(i)}$, the test subject is matched to the sample that satisfies the following:
\begin{gather*}
i^*=\argmin_i \big[ dis(F^*,  F^{(i)}) \big]
\end{gather*}
 Here, each feature matrix will have a size of $14\times64$ due to the size of the images and the blocks. $M$ of the 12 samples from every person are assigned as training and the rest as test cases, adding up to $500(12-M)$ test subjects. The feature matrix is defined as the average of the feature matrices of the $M$ training images. For an unknown sample with the feature matrix $F^*$, the following distance will be:
\begin{gather*}
dis(F^*,F^{(i)})= \sum_{m=1}^{14} \sum_{n=1}^{64} w_m \alpha_m (F^*_{mn}-F^{(i)}_{mn})^2
\end{gather*}
Each row has a weight of $w_m\alpha_m$, where $\alpha_m$ is a feature normalizing factor trying to map all features into the same range and is defined as the reciprocal of the mean value of the corresponding feature of all training samples, while $w_m$ is the feature importance factor which is higher as the usefulness of the feature increases.  Here $w_m$ is defined as the recognition accuracy when the $m$-th row of the feature matrix is used for recognition.
We should find the distance defined earlier for all the spectra by comparing the images in the same color. Next, the distance between a test image and the $i$-th training sample will be defined as the average of the distances of their corresponding spectra. In the end, the prediction for a test image with the feature matrix $F^*$ is:
\begin{gather*}
i^*=\argmin_i \big[ dis(F^*,  F^{(i)}) \big]
\end{gather*}

\section{Results}
\label{SectionIV}

In our dataset, each image is preprocessed and aligned and has the resolution of 128$\times$128. In each setting of our experiment, we have performed recognition for various combinations of training data and test subjects. Whenever the test data does not match our expectation, it is an ID fail or misidentification.

 The results from majority voting and minimum distance classifications are shown in Table \ref{TblRes}. For majority voting, due to the much shorter time it takes, every test is repeated 10 times and their average is recorded. For the minimum distance classifier, the image permutations are adjacent. For example, two neighbor minimum distance cases are common in all their training data selections but one.
 
\begin{table} [h]
  \caption{Accuracy rates of minimum distance classifier and weighted majority voting algorithm}
  \centering
\begin{tabular}{|m{1cm}|m{1.3cm}|m{1.3cm}|m{1.3cm}|m{1.3cm}|}
\hline
\multirow{2}{*}{\parbox{1cm}{Training sample fraction}} & \multicolumn{2}{c}{Minimum Distance} & \multicolumn{2}{|c|}{Majority Voting}  \\
\cline{2-5}
 & No feature weight & Weighted features & No feature weight & Weighted features \\
\hline
\ \ 4/12 & \ \ \ 96.45 & \ \ \ 96.35 &  \ \ \ 99.96 & \  \ \ 99.96  \\
\hline
\ \ 5/12 & \ \ \ 95.71 & \ \ \ 97.46 & \ \ \ 100 & \ \ \ 99.99   \\
\hline
 \ \ 6/12  & \ \ \ 98.07 & \ \ \ 97.80 & \ \ \ 100 & \ \ \ 100   \\
\hline
 \ \ 7/12 & \ \ \ 97.40 & \ \ \ 96.92 & \ \ \  100 & \ \ \  100   \\
\hline
 \ \ 8/12  & \  \ \ 98.65 & \ \ \ 98.35 & \ \ \  99.99  & \ \ \ 99.99 \\
\hline
  \ \ 9/12  & \ \ \ 97.60 & \ \ \ 97.33 & \ \ \ 100  & \ \ \ 100  \\
\hline
 \ 10/12  &  \ \ \ 98.60 & \ \ \ 98.40 & \ \ \ 100  & \ \ \ 100  \\
\hline
\end{tabular}
\label{TblRes}
\end{table}
Table \ref{TblRes} shows that the performance of the majority voting classifier is much more efficient than the minimum distance classifier. 

Table \ref{TblComp} shows a comparison of the results of our work and those of three other accurate and relatively newer algorithms. We compared the results for three different train-to-test ratios as the others also reported these three cases. Note that the blank spaces under QPCA are due to them being not reported in the source.

\begin{table} [h]
\centering
  \caption{Comparison with other algorithms for palmprint recognition }
  \centering
\begin{tabular}{|m{1.2cm}|m{1.1 cm}|m{1.1cm}|m{1.1cm}|m{1.1cm}|}
\hline
Training sample fraction & QPCA \cite{10} & Hybrid \ \ \ \ \ \ \ feature \cite{4} & Stat/Wave* \ \ \ \cite{9} (MDC) & Proposed method (WMV)\\
\hline
\ \ \ 6/12 & \ \ 98.13\% & \ \ 98.88\% & \ \ 100\% &  \ \ 100\%\\
\hline
\ \ \ 5/12 & \ \ - & \ \ 98.45\%  & \ \ 99.77\% & \ \ 99.99\% \\
\hline
\ \ \ 4/12 &  \ \ - & \ \ 98.08\% & \ \ 99.65\% & \ \ 99.96\%\\
\hline
\end{tabular}
*Statistical and wavelet features
\label{TblComp}
\end{table}

As it can be seen, our algorithm has a higher accuracy rate compared to the previous works and also slightly outperforms the results from \cite{9}. A comparison between our work and some of the others is illustrated in Figure 2. 
\begin{figure}[2 h]
\begin{center}
    \includegraphics [scale=0.5] {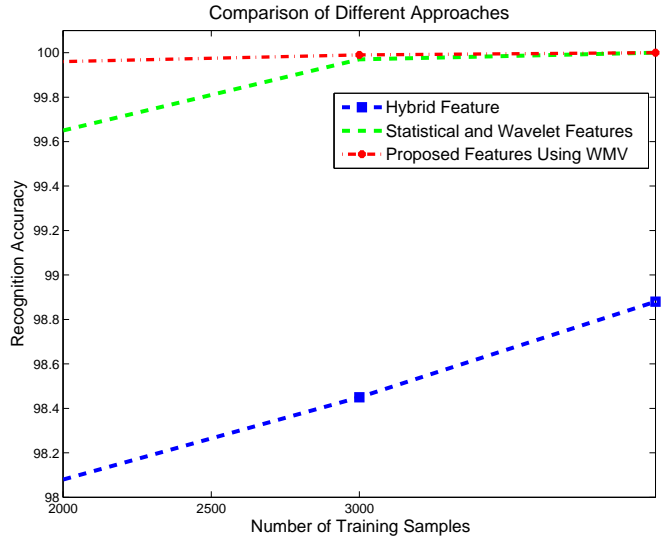}
\end{center}
  \caption{Comparison of different palmprint recognition approaches}
\end{figure}

The method is tested using MATLAB on a laptop with Windows 7 and Core i7 running at 2GHz. The calculation time for weighted majority voting is 0.06s per test, while minimum distance classifier takes 0.09s per test.

\section*{Conclusion}
\label{SectionV}

This paper proposed a set of textural features based on co-occurrence for palmprint recognition. This method senses the textures of the images and extracts 14 features from them. Two different classifiers, weighted majority voting and minimum distance classifiers, are also used to perform the recognition.
The proposed scheme has advantages over many older popular methods. It has a very high accuracy rate as well as a low processing time, making it possible to use in real-time applications. The calculation of the features are also straightforward. There are many speculations for the future including applying the same features to other biometrics such as fingerprints and iris patterns.

\section*{Acknowledgments}
The authors of this paper would like to express their gratitude to the Hong Kong Polytechnic University (PolyU) for sharing their invaluable palmprint database without which this project could not be completed.



\begin{thebibliography}{1}

\bibitem{1}
 A. K. Hrechak and J. A. McHugh, ``Automated fingerprint recognition using structural matching,'' Pattern Recognition 23.8 (1990): 893-904.
\bibitem{2}
D. M. Monro, S. Rakshit and D. Zhang, ``DCT-based iris recognition,'' Pattern Analysis and Machine Intelligence, IEEE Transactions on 29.4 (2007): 586-595.
\bibitem{3}
W. Zhao, R. Chellappa, P. J. Phillips and A. Rosenfeld, ``Face recognition: A literature survey,'' ACM Computing Surveys (CSUR) 35, no. 4 (2003): 399-458.
\bibitem{4}
S. A. Mistani, S. Minaee and E. Fatemizadeh, ``Multispectral palmprint recognition using a hybrid feature,'' arXiv preprint arXiv:1112.5997 (2011).
\bibitem{5}
X. Wu, K. Wang and D. Zhang, ``Wavelet energy feature extraction and matching for palmprint recognition,'' Journal of Computer Science and Technology 20.3 (2005): 411-418.
\bibitem{6}
W. K. Kong, D. Zhang and W. Li, ``Palmprint feature extraction using 2-D Gabor filters,'' Pattern recognition 36.10 (2003): 2339-2347.
\bibitem{7}
J. Chen, C. Zhang and G. Rong, ``Palmprint recognition using crease,'' Proceedings, 2001 International Conference on Image Processing, Vol. 3. IEEE, 2001.
\bibitem{8}
W. Jia, D. Huang and D. Zhang, ``Palmprint verification based on robust line orientation code,'' Pattern Recognition 41.5 (2008): 1504-1513.
\bibitem{9}
S. Minaee and A. Abdolrashidi, ``Highly Accurate Multispectral Palmprint Recognition Using Statistical and Wavelet Features,'' arXiv preprint arXiv:1408.3772 (2014).
\bibitem{10}
X. Xu and Z. Guo, ``Multispectral palmprint recognition using quaternion principal component analysis,'' IEEE Workshop on Emerging Techniques and Challenges for Hand-Based Biometrics, pp. 1–5, 2010.
\bibitem{Database}
http://www.comp.polyu.edu.hk/\~biometrics/MultispectralPalmprint/MSP.htm
\bibitem{12}
D. Zhang et al., ``An online system of multispectral palmprint verification,'' Instrumentation and Measurement, IEEE Transactions on 59.2 (2010): 480-490.
\bibitem{13}
S. Minaee, M. Fotouhi and B. H. Khalaj, ``A geometric approach for fully automatic chromosome segmentation,'' arXiv preprint arXiv:1112.4164 (2011).
\bibitem{14} 
R. M. Haralick, K. Shanmugam and I. Dinstein, ``Textural features for image classification,''  IEEE Transactions on Systems, Man and Cybernetics, no. 6 (1973): 610-621.
\bibitem{15}
S. Minaee, Y. Wang and Y. W. Lui, ``Prediction of Longterm Outcome of Neuropsychological Tests of MTBI Patients Using Imaging Features,'' Signal Processing in Medicine and Biology Symposium (SPMB), IEEE, 2013.

\end{thebibliography}
\end{document}